\documentclass[10pt,twocolumn,letterpaper]{article}

\usepackage{cvpr}
\usepackage{times}
\usepackage{epsfig}
\usepackage{graphicx}
\usepackage{amsmath}
\usepackage{amssymb}
\usepackage{breqn}
\usepackage{booktabs} 
\usepackage{color}
\usepackage{tabularx}
\usepackage{mathrsfs}
\usepackage{multirow}

\definecolor{darkgreen}{RGB}{10,150,10}

\usepackage[pagebackref=true,breaklinks=true,letterpaper=true,colorlinks,bookmarks=false,citecolor=blue,linkcolor=blue]{hyperref}

\cvprfinalcopy 

\def\cvprPaperID{1612} 

\ifcvprfinal\fi
\begin{document}

\title{Controllable and Progressive Image Extrapolation}

\author{Yijun Li$^{1,2}$ ~~~~~~~Lu Jiang$^3$ ~~~~~~Ming-Hsuan Yang$^{1,3}$\\
$^1$UC Merced~~~~~~~~~~~$^2$Adobe Research~~~~~~~~~~$^3$Google Research\\
{\tt\small yijli@adobe.com ~~~~~lujiang@google.com~~~~~~mhyang@ucmerced.edu}
}

\maketitle

\begin{abstract}

Image extrapolation aims at expanding the narrow field of view of a given image patch. 
Existing models mainly deal with natural scene images of homogeneous regions and have no control of the content generation process.
%
In this work, we study conditional image extrapolation to synthesize new images guided by the input structured text.
The text is represented as a graph to specify the objects and their spatial relation to the unknown regions of the image.
%
Inspired by drawing techniques, we propose a progressive generative model of three stages, i.e., generating a coarse bounding-boxes layout, refining it to a finer segmentation layout, and  mapping the layout to a realistic output. 
Such a multi-stage design is shown to facilitate the training process and generate more controllable results.
We validate the effectiveness of the proposed method on the face and human clothing dataset in terms of visual results, quantitative evaluations and flexible controls.

\end{abstract}


\section{Introduction}


Given an image patch with a narrow field of view, image extrapolation aims at expanding it by generating plausible visual content outside the image boundaries.
The extrapolation is a challenging task since it requires to synthesize new content that aligns well with the given image patch. 
%
To the best of our knowledge, only a few  approaches~\cite{shan-2014-photo,zhang-2013-framebreak,wang-2014-biggerpicture}
have been developed to address this topic, and all are designed for \emph{unconditional} extrapolation where the target image is generated solely based on the input patch. 
This is often achieved by finding low-level cues of similar patterns from the given image or external databases. 
These methods perform well on natural images of homogeneous regions.

\begin{figure}[t]
\centering
\begin{tabular}{c@{\hspace{0.01\linewidth}}c@{\hspace{0.01\linewidth}}c@{\hspace{0.01\linewidth}}c@{\hspace{0.01\linewidth}}c@{\hspace{0.01\linewidth}}c}


\includegraphics[height = .5\linewidth]{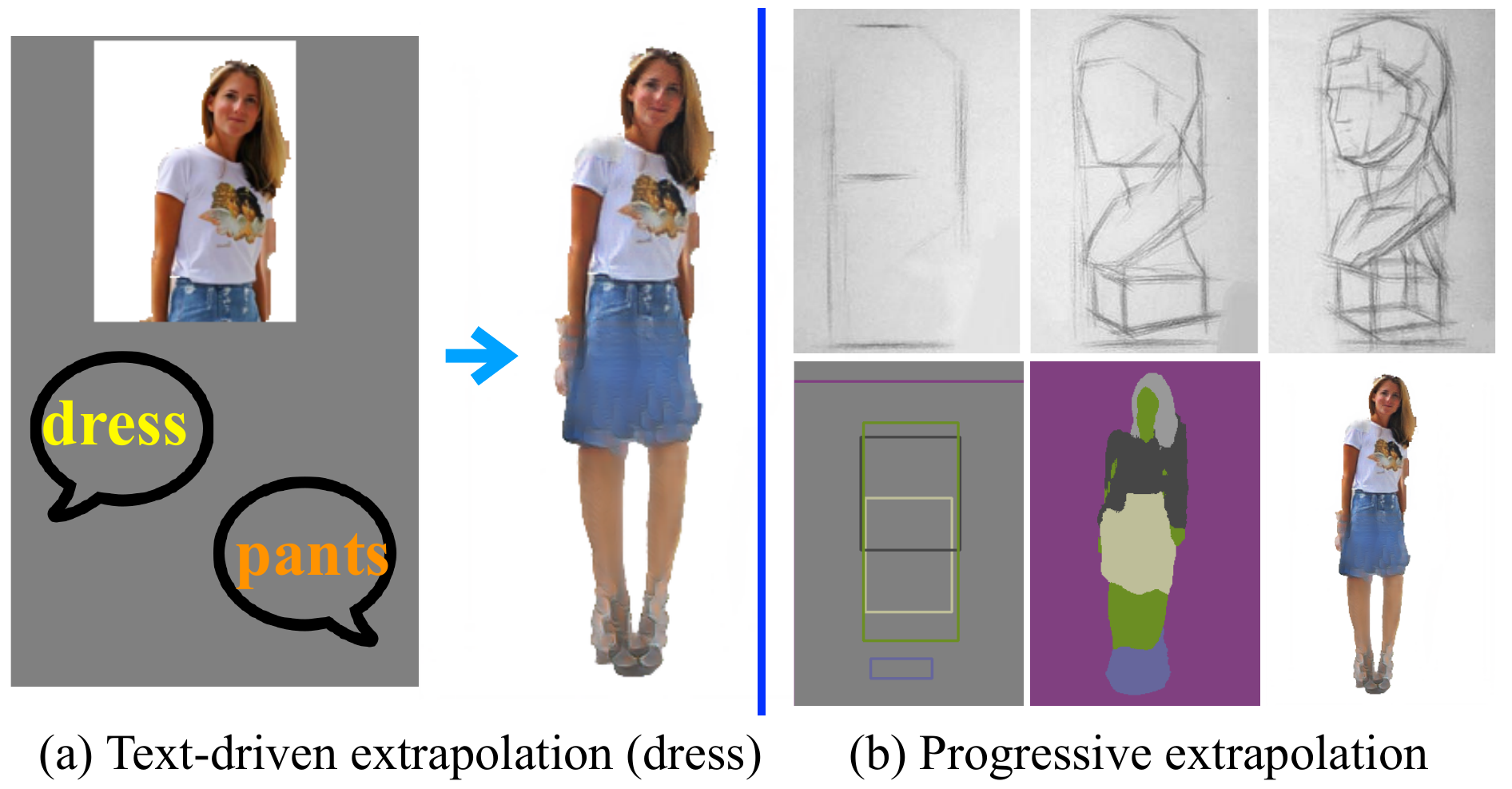} &







\end{tabular}

\caption{Definition and motivation of the extrapolation task.
(a) Conditional image extrapolation takes the input of the image patch and text. 
Users may want to synthesize the lower body  to generate the \emph{dresses} or \emph{pants} object and can control the generation by the text input.
(b) Top: illustration of human layout drawing in the coarse-to-fine manner.
Bottom: intermediate and final outputs of our progressive generation model, which corresponds to each step of human layout drawing.
}
\label{fig:motivation}
\vspace{-1.5em}
\end{figure}

A core problem, however, is that oftentimes a user has some concept in mind from which one wants to generate an image, and the most straightforward way to express the concept is via text.
Consider an example in Figure~\ref{fig:motivation}(a), for the given patch, users may have different ideas of extrapolating the lower body, wearing the dress or pants.
An ideal model should directly take both the patch and text into account to generate the target image. 
%

In this paper, we study \emph{conditional image extrapolation} where the inputs are an image patch and a structured text that specifies desired properties to synthesize. 
The image patch serves as the same role as that in the unconditional extrapolation, whereas the input text controls the content generation outside the image boundaries. 
%
Similar to~\cite{johnson-2018-sg2im}, we represent the structured text as a scene graph to circumvent handling the ambiguity in natural languages.
The scene graph~\cite{lu2016visual,xu2017scene,woo2018linknet,johnson-2018-sg2im} consists of nodes to represent objects and edges to describe their relations (spatial arrangements in our case). 
Conditional image extrapolation offers more flexibility than existing counterparts in that users can \emph{control} what and where to generate outside the image boundaries, thereby allowing users to generate a variety of target images from the same image patch with different text descriptions.
%
%
%
Our problem is related to text-to-image generation~\cite{zhang-2017-stackgan,zhang-2017-stackgan++,yan-2016-attribute2image} but differs in its usage of multimodal input of both image and text. 



A straightforward solution to this problem is to learn a deep generative model (\eg,~\cite{isola-2017-pix2pix,wang-2018-pix2pixhd,park-2019-SPADE,chen2017photographic}) to directly translate unknown regions to plausible RGB pixels. 
However, this approach is likely to generate blurry images of poor quality. More importantly the text cannot effectively control the generated content.
The reason is that learning such a direct mapping between two different modalities (from text to high-dimensional pixel space) is extremely difficult. 
As a result, the current key research question for conditional image extrapolation is how to make the image generation process controllable by the input text and amicable to the input image patch.

To address this issue, we mimic the 
process of how an painter creates an artwork.
Before filling out the details, a painter often progressively refine a sketch from object contours to finer layouts, as shown on the top row of Figure~\ref{fig:motivation}(b).
%
Motivated by this, we propose a progressive generative model that consists of three stages to extrapolate an image patch.
%
%
We first generate a \emph{bounding-box layout} from the scene graph to roughly indicate the size and spatial location of each object.
Conditioned on the bounding-box layout, we then learn to generate a semantic \emph{segmentation layout}, where each pixel is represented as an object class label. 
Finally, we map the segmentation layout to the extrapolated pixels via image-to-image translation. See the bottom row of Figure~\ref{fig:motivation}(b).
These modules are first separately trained for individual tasks and then jointly optimized.

%
%
%
%

We evaluate the conditional image extrapolation on two public datasets in terms of visual results, quantitative evaluations and flexible controls. 
Extensive experimental results demonstrate that our model performs favorably against existing methods.
The progressive training not only speeds up the convergence substantially but also makes the generated content more controllable. 
%
In addition, the intermediate outputs, byproducts of our model, are semantically meaningful to users.
%
%
%
%
The main contributions of this work are summarized as follows:
\begin{itemize} 
\item We study a new task of conditional image extrapolation which takes multimodal inputs of image and text.
\item We propose an effective progressive generative network to synthesize new content outside image boundaries by generating layouts as sub-tasks.
\item We realize controllable extrapolation to generate diverse extrapolated images which respect different indications in the scene graph.
\end{itemize}

\section{Related Work}
\label{sec:related}


\noindent\textbf{Image extrapolation.}~Early extrapolation algorithms generally follow a retrieve-and-compose strategy where an external library of sample images that depict the similar scene is assumed to be available.
For example, Efros and Freeman~\cite{efros2001image} expand the small texture patch with similar patches and develop an optimal boundary with minimum cost for composition.
By extending similar textured patches to images of the similar scene category, Zhang \etal~\cite{zhang-2013-framebreak} extrapolate photos by utilizing the self-similarity of a reference image to generate a set of local transformations.
To handle different viewpoints and appearance variations, a few methods~\cite{shan-2014-photo,wang-2014-biggerpicture} use library images to search good candidates and align them with the given input.
However, those non-parametric methods are mainly limited in inferring semantically new content and requiring proper reference databases. 

\vspace{0.5em}
\noindent\textbf{Conditional image generation.}~With the recent advances of generative models~\cite{GAN-2014-generative,cGAN-2014-conditional}, filling the unknown regions in the image is categorized as the conditional image generation problem. 
For texture synthesis, Zhou \etal \cite{zhou-2018-non} directly train a feed-forward network to expand a certain small texture patch to a larger one. 
For photos, a number of image inpainting methods~\cite{GFC-CVPR-2017,yu-2018-generative,iizuka2017globally,liu-2018-image,wang-2018-image} learn to fill the holes inside the image with different design of architectures and losses, and achieve better results over  diffusion-based~\cite{bertalmio2000image,sun-2005-image} or patch-based~\cite{barnes-2009-patchmatch} schemes.
However, those approaches seldom pay attention to image extrapolation where the number of unknown pixels is much more than that of known pixels, and lack controls to generate diverse results. 
Two recent methods are developed to 
to extrapolate images but still under the uncontrollable setting \cite{wang-2019-IE,yang2019very}.

\begin{figure*}[t]
\centering
\begin{tabular}{c@{\hspace{0.01\linewidth}}c@{\hspace{0.01\linewidth}}c@{\hspace{0.01\linewidth}}c@{\hspace{0.01\linewidth}}c@{\hspace{0.01\linewidth}}c}

\includegraphics[height = .215\linewidth]{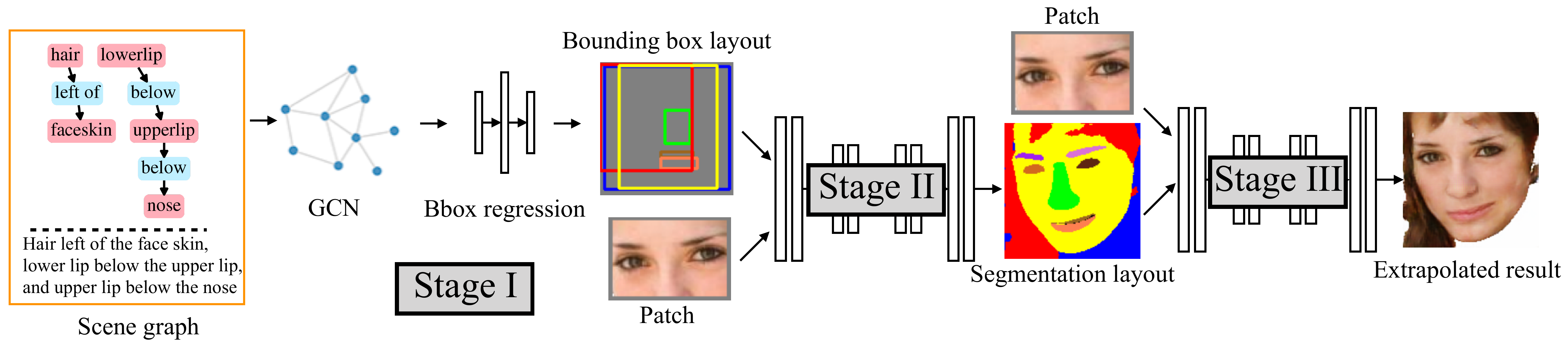} & \\


\end{tabular}
\caption{Framework of the proposed algorithm on progressive extrapolation.
In stage I, we generate a bounding-box layout from the scene graph to roughly indicate the size and spatial location of each object. 
Then conditioned on the coarse bounding-box layout and the image patch, we learn to generate a semantic segmentation layout in stage II. 
Finally in stage III, we map the segmentation layout and the image patch to generate the extrapolated results. Details about the network architecture can be found in the supplementary material.}
\label{fig:framework}
\vspace{-1em}
\end{figure*}

Our work is related to image generation controlled by other signals such as a reference image~\cite{MUNIT-2018-multimodal,DRIT-2018-diverse}, a label~\cite{cGAN-2014-conditional,stargan-2018}, or a sentence~\cite{yan-2016-attribute2image,zhang-2017-stackgan,zhang-2017-stackgan++,zhang-2018-photographic} (also known as text-to-image generation).
Different from existing approaches, the proposed method takes the input of both image and text in which the unique challenge lies in aligning the generated image with the input patch guided by the text.
%
Among all recent methods, closest to ours is the work by Johnson~\etal~\cite{johnson-2018-sg2im} which uses scene graphs to generate images. 
In contrast, our work is different in two aspects: (i) the multimodal input (ii) the  progressive training that enables controllable image extrapolation. 
%


\vspace{0.5em}
\noindent\textbf{Layout generation.}~Recently, numerous methods have been developed for inferring a reasonable layout from an image. 
Some inpainting approaches \cite{xiong2019foreground,nazeri2019edgeconnect} recover the missing edges to complete the structures first before generating the final results. 
Li \etal~\cite{li-2019-layoutganjm} propose a generative adversarial network (GAN) to learn the distribution of layout for graphic design and generation. 
%
%
Different from these methods that infer silhouettes of each image category, the proposed model focuses more on extrapolating plausible parts and layouts of object images. 

\vspace{0.5em}
\noindent\textbf{Curriculum and progressive learning}
%
Our progressive training approach is related to curriculum learning schemes ~\cite{bengio2009curriculum}, which aim to master a complex job by first learning easier aspect of the task and gradually take more complex samples or sub-tasks into consideration. 
It has been widely used to order/weight training samples~\cite{jiang2018mentornet,chang2017active} or to prioritize the tasks in multi-task learning~\cite{sharma2017learning,pentina2015curriculum}. 
Furthermore, a number of recently proposed progressive methods~\cite{karras2017progressive,liu2018progressive} share the similar underlying idea.
%
The line of research work on curriculum learning generally regards finding an optimal order of executing some known tasks~\cite{bengio2009curriculum}. 
Different from prior work, the sub-tasks in our problem, \ie, the bounding-box and segmentation layout, are unknown. 
Our work designs two latent tasks and empirically demonstrates their efficacy for conditional image extrapolation.

\section{Proposed Method}

Given an input image patch and a structured text represented as a scene graph, our goal is to extrapolate visual content beyond image boundaries that satisfies the conditions specified in the scene graph. 
We formulate this problem as a conditional image generation problem, where the conditions are the image patch, which specifies visual content in the known region of the target image, and the text (scene graph) which defines desired objects and their spatial relation to extrapolate for the unknown region.

Our model takes two inputs: an image patch $\mathbf{z}_p$ and a structured text represented as a scene graph $\mathbf{sg}$. 
We denote the input image patch as $\mathbf{z}_p \in \mathbb{R}^{h \times w \times 3}$ and the target image to generate as $\mathbf{x} \in \mathbb{R}^{H \times W \times 3}$, where $h,H$ and $w,W$ are width and height of the images and $h < H$, $w < W$. 
We represent the text input as a scene graph~\cite{johnson2015image}. 
Given a set of pre-specified object categories $\mathcal{C}$ and relationship categories $\mathcal{R}$, a scene graph is a tuple $\mathbf{sg} = (O, E)$ where $O = \{o_i | o_i \in \mathcal{C}\}$ is a set of objects to extrapolate for the unknown region, and $E \subseteq O \times \mathcal{R} \times O$ is a set of directed edges specifying the relationship between objects.
We focus on a common type of relationship in our problem, \ie, the spatial relationship between objects which includes $\{$left of, right of, above, below, inside, surrounding$\}$.

Given an training example $\mathbf{x}$ drawn from the real distribution $p_{real}$ and $\mathbf{z}_p$ randomly cropped from $\mathbf{x}$, our generation model learns a mapping function from $\mathbf{z}_p$ and $\mathbf{sg}$ to the data space $\hat{\mathbf{x}}=G(\mathbf{z}_p , \mathbf{sg} ;\theta_g) \in \mathbb{R}^{H \times W \times 3}$. 
In general, this learning process is self-supervised with a reconstruction loss $L_{rec}$ and an adversarial loss $L_{adv}$~\cite{GAN-2014-generative}:
\begin{equation}
    L_{total} = L_{rec} + \lambda L_{adv} = \mid \mid \mathbf{x} - \hat{\mathbf{x}} \mid \mid^{2}_{2} + \lambda L_{adv},
\label{eq:gan_objective}
\end{equation}
%
%
where $L_{adv}$ is computed by:
\begin{equation}
    L_{adv} \!=\! \mathbb{E}_{\mathbf{x} \sim p_{real}} [\log D(\mathbf{x})]
     +  \mathbb{E}_{\mathbf{\hat{x}} \sim p_{fake}} [\log(1-D(\mathbf{\hat{x}}))],
\label{eq:gan_objective1}
\end{equation}
where $D$ is a discriminator to output a single scalar representing the probability of whether the $\mathbf{x}$ is real or not.

\begin{figure}[t]
\centering
\begin{tabular}{c@{\hspace{0.01\linewidth}}c@{\hspace{0.01\linewidth}}c@{\hspace{0.01\linewidth}}c@{\hspace{0.01\linewidth}}c@{\hspace{0.01\linewidth}}c}

\includegraphics[height = .33\linewidth]{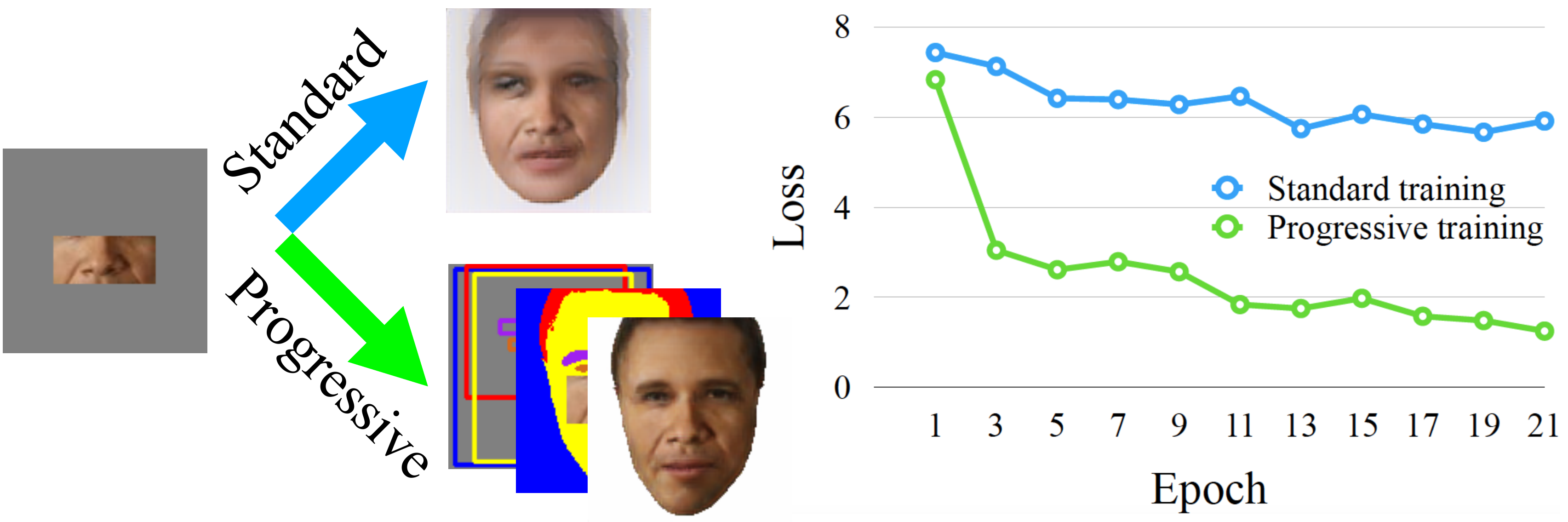} & \\


\end{tabular}
\caption{Comparison of the standard training and the proposed progressive training.
}
\label{fig:onenetwork}
\vspace{-1em}
\end{figure}

\subsection{Overview}

%
Directly optimizing Eq.~\eqref{eq:gan_objective} with deep generation networks (\eg,~\cite{isola-2017-pix2pix,wang-2018-pix2pixhd,park-2019-SPADE,chen2017photographic}) to translate unknown regions to plausible regions (\ie, the standard training) only leads to blurry and less realistic outputs. 
Figure~\ref{fig:onenetwork} shows an example training curve where the training loss (in blue) hardly decreases after a few epochs. 
The underlying reason is that using text to directly control RGB pixel generation is extremely difficult.

To address this issue, we design two latent sub-tasks that are closely related to our final generation task but are progressively easier to learn.
%
%
Specifically, we train the generator progressively via three tasks where the output of a previous task is used in the next task. Let $\theta_g^*$ be the optimal parameter for our generator $G$ and we find it by minimizing the total loss $L_{total}$ over all training pairs of scene graphs and image patches:
%
\begin{equation}
\begin{split}
    \label{eq:decompose_generation}
    L_{total}
    = L_{box}(\mathbf{sg}) + L_{seg} (\mathbf{x}_{bb}, \mathbf{z}_{p}) + L_{img} (\mathbf{x}_{seg}, \mathbf{z}_{p}),
\end{split}
\end{equation}
where the losses $L_{box}, L_{seg}$, and $L_{img}$ are used to estimate the negative log-likelihood for each generation of $p(x_{bb} | sg)$, $p(x_{seg} | x_{bb}, z_{p})$, and $p(\hat{x} | x_{seg}, z_{p})$, respectively. 
With Eq.~\eqref{eq:decompose_generation}, 
the generation process is decomposed into three stages. 
First the bounding-box layout $\mathbf{x}_{bb}$ is constructed from the scene graph. Then the segmentation layout $\mathbf{x}_{seg}$ is created from the bounding-box and the input patch. 
Finally, the model generates the target image  $\mathbf{\hat{x}}$ using the segmentation layout and the input patch.


Figure~\ref{fig:framework} illustrates the framework of our model.
Our network first generates a \emph{bounding-box layout} $\mathbf{x}_{bb} \in \mathbb{Z}^{|O| \times 4}$, a low-dimensional coordinate space for each object in the scene graph. 
Then the bounding-boxes are refined into a semantic \emph{segmentation layout} ($\mathbf{x}_{seg} \in \mathbb{Z}^{H \times W \times 1}$), where each pixel is represented as a classification label of the object in $O$. 
The third stage maps the segmentation layout to the extrapolated RGB pixels $\mathbf{\hat{x}}$ via image-to-image translation. 
%
In the following, we describe the details of these stages. 

%
%
%
%
%


\subsection{Stage I: Bounding-box Layout Generation}\label{sec:stage-i}

\begin{figure}[t]
\centering
\begin{tabular}{c@{\hspace{0.005\linewidth}}c@{\hspace{0.005\linewidth}}c@{\hspace{0.005\linewidth}}c@{\hspace{0.005\linewidth}}c@{\hspace{0.005\linewidth}}c@{\hspace{0.005\linewidth}}c@{\hspace{0.005\linewidth}}c@{\hspace{0.005\linewidth}}c}

\includegraphics[width = .96\linewidth]{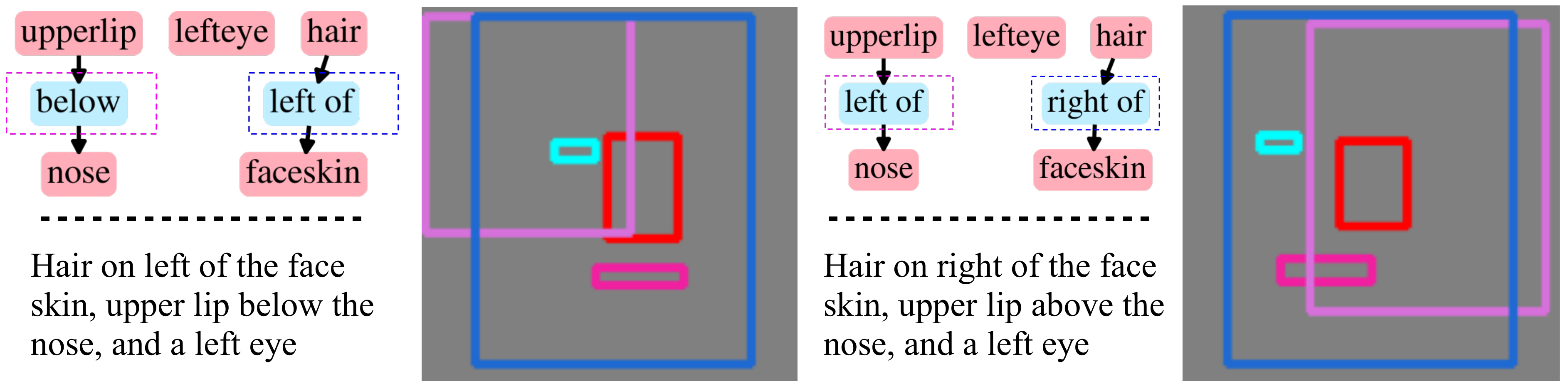} & \\
{(a) Stage I: bounding-box generation } \\

\includegraphics[width = .96\linewidth]{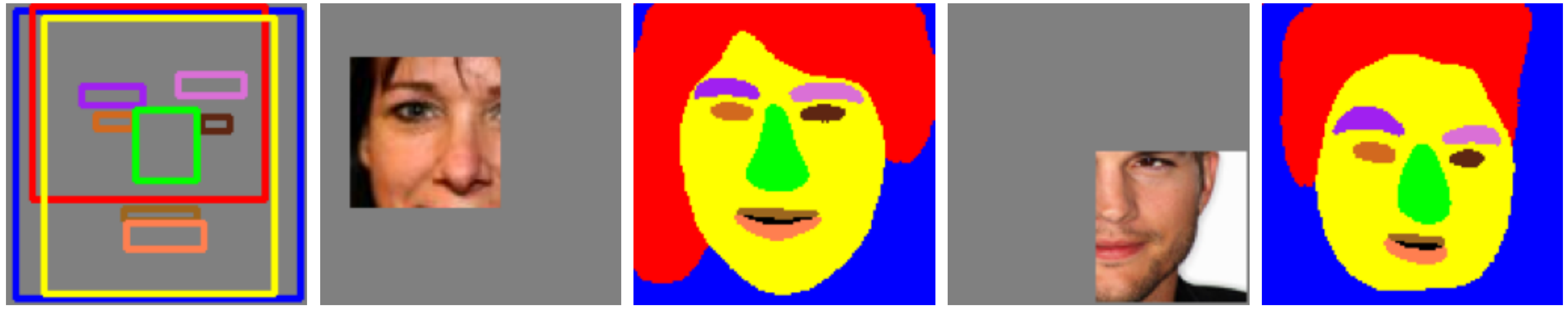} & \\

{(b) Stage II: segmentation layout generation } \\

\end{tabular}
\caption{Examples of outputs of Stage I and II.}
\label{fig:stage2}
\vspace{-1em}
\end{figure}

%
%
The Stage I takes the scene graph as input and outputs a bounding-box spatial layout map. For the scene graph input, we use the graph convolution network (GCN) of~\cite{johnson-2018-sg2im} to transform object embeddings into the relationship-encoded representation.
Given a graph with embeddings initialized at each
node and edge, the GCN computes new embeddings for each node and edge through propagating information along edges of the graph. 
The edge embedding encodes the relationship between connected objects.
%
%
%
The encoded object embeddings are then fed into a fully-connected network of three layers to predict the bounding-box coordinates $\hat{bb}_i$ for each object. 
Each box is represented as the top-left and bottom-right $x$-$y$ coordinates. 
%
%
The loss in this stage is computed by the $L_{1}$ difference between ground-truth and predicted boxes:
\begin{equation}
    L_{box} = \frac{1} {|O|}\sum^{|O|}_{i=1} \mid \mid bb_{i} - \hat{bb}_{i}\mid \mid_{1}
\end{equation}
where $bb_{i}$ is the true bounding-box.
Figure~\ref{fig:stage2}(a) shows two examples of the generated bounding-box layout for different scene graph inputs.
%
%

Note that sometimes scene graphs can be similar, \eg, nearly all face images contain ``eyes'' and ``nose'' as the nodes. 
The lack of diversity makes it difficult to learn a good graph embedding.
To address this issue, we augment the training data by randomly dropping some nodes out of the scene graph and meanwhile modifying the target image accordingly. 
%
%
We observe that the augmentation considerably enhances the controllability of the scene graph. 
%

\vspace{0.5em}
\subsection{Stage II: Segmentation Layout Generation}
The Stage II is responsible for transforming the coarse bounding-box layout into a segmentation layout conditioned on the image patch. 
As such, we need to accomplish three goals: (i) parse the known regions in the patch, (ii) generate the segmentation layout for the unknown regions, and (iii) align the unknown and known regions.

The input to Stage II is the concatenated feature of the graph embedding from Stage I and the input image patch. 
We warp each node embedding in the scene graph using bilinear interpolation according to coordinates to 
compute a spatial vector that has the same shape as the input image.
%
%
We use the network of~\cite{pohlen2017-frrn} as the backbone architecture to infer the pixel-level object labels. 
Let $c_1$, \ldots, $c_N$ $\in$ \{1, \ldots, $|\mathcal{C}|$\} be the target class labels for the pixels $1, \cdots, H \times W$ where $|\mathcal{C}|$ is the number of object categories and $N= H \times W$. 
This module is trained with pixel-wise multi-class cross-entropy loss:
\begin{equation}
\label{eq:cross_entropy}
    L_{seg} = -\frac{1}{N} \sum_{i=1}^{N} \sum_{c_i=1}^{|\mathcal{C}|} \omega_{c_{i}} ~ y_{i, c_{i}} \log p_{i, c_{i}},
\end{equation}
where $p_{i, c_{i}}$ is the predicted probability for pixel $i$ of belonging to class $c_i$, and $y_{i, c_{i}}$ is the binary label (0 or 1) indicating if class label $c_{i}$ is a correct classification for pixel $i$. To handle the imbalanced classes (\eg, ``background'' class is more common than ``eye''), we use $\omega_{c_{i}}$ to downweigh the pixels from common classes.

Figure~\ref{fig:stage2}(b) shows two examples of the alignment between an existing eye in the given patch and the other eye generated outside boundaries.
Given the same bounding-box layout, while the eyes of two conditional patches are at different height, our model is able to generate different segmentation layout that aligns with the input image patch well.
This indicates that bounding-box layouts only impose soft constraints, and Stage II is able to recover the error from the Stage I output. 

\vspace{0.5em}
\subsection{Stage III: Layout to Image Generation}
Given the generated layout, the Stage III operates as a label-to-image mapping model in a way similar to image-to-image translation~\cite{isola-2017-pix2pix,park-2019-SPADE}. 
Here we use a generic auto-encoder with the instance normalization layer~\cite{ulyanov2017-improvedin} for regularizing the network activations.
The difference to image-to-image translation here is that our input is the concatenation of the segmentation layout and the input image patch.
%
%
To learn this model, we use the perceptual loss~\cite{johnson2016perceptual} and adversarial loss~\cite{arjovsky2017-wgan}:
\begin{equation}
L_{img} = \sum \limits_{i=1}^{4} \mid \mid \Phi_{i}(\mathbf{x}) - \Phi_{i}(\mathbf{\hat{x}}) \mid \mid^{2}_{2} + L_{adv}~,
\end{equation}
where $\mathbf{x}$, $\mathbf{\hat{x}}$ are the ground truth and predicted image, and $\Phi_{i}$ is the pretrained VGG-19~\cite{VGG-2014} network up to the ReLU\_$i$\_1 layer. 
%

\vspace{0.5em}
\noindent\textbf{Remarks on training.}~While all three tasks share the same goal of extrapolating valid objects that align well with the given image patch, they are made increasingly difficult to learn. 
For example, it is much easier to find the box locations (in Stage I) than the RGB image (in Stage III) for all objects to satisfy their relationship in the scene graph. 
Likewise, it is a simpler task to align the input image patch with the boxes than the final extrapolated content.
%
%
%
%
%
%
Therefore, we first train each stage separately such that each stage can focus on its own objective and learn a better initialized model than random weights. 
However, the individual models trained in these stages 
may cause errors when the intermediate stage does not generate the precise layout.
We further jointly train all three models (from three stages) in order to enforce the later stage to correct some inconsistent outputs from the previous stage. 
%
%
%

\section{Experiments}
\label{sec:exp}
We conduct experiments to validate the effectiveness of our model for conditional image extrapolation.
We evaluate the proposed method on two types of object images of great interests, \ie, face and human body, on two public benchmark datasets against existing algorithms.
The source code and trained models will be made available to the public.
More results and details can be found in the supplementary material.

\begin{figure*}[t]
\centering
\begin{tabular}{c@{\hspace{0.005\linewidth}}c@{\hspace{0.005\linewidth}}c@{\hspace{0.005\linewidth}}c@{\hspace{0.005\linewidth}}c@{\hspace{0.005\linewidth}}c@{\hspace{0.005\linewidth}}c@{\hspace{0.005\linewidth}}c@{\hspace{0.005\linewidth}}c}

\includegraphics[width = .27\linewidth]{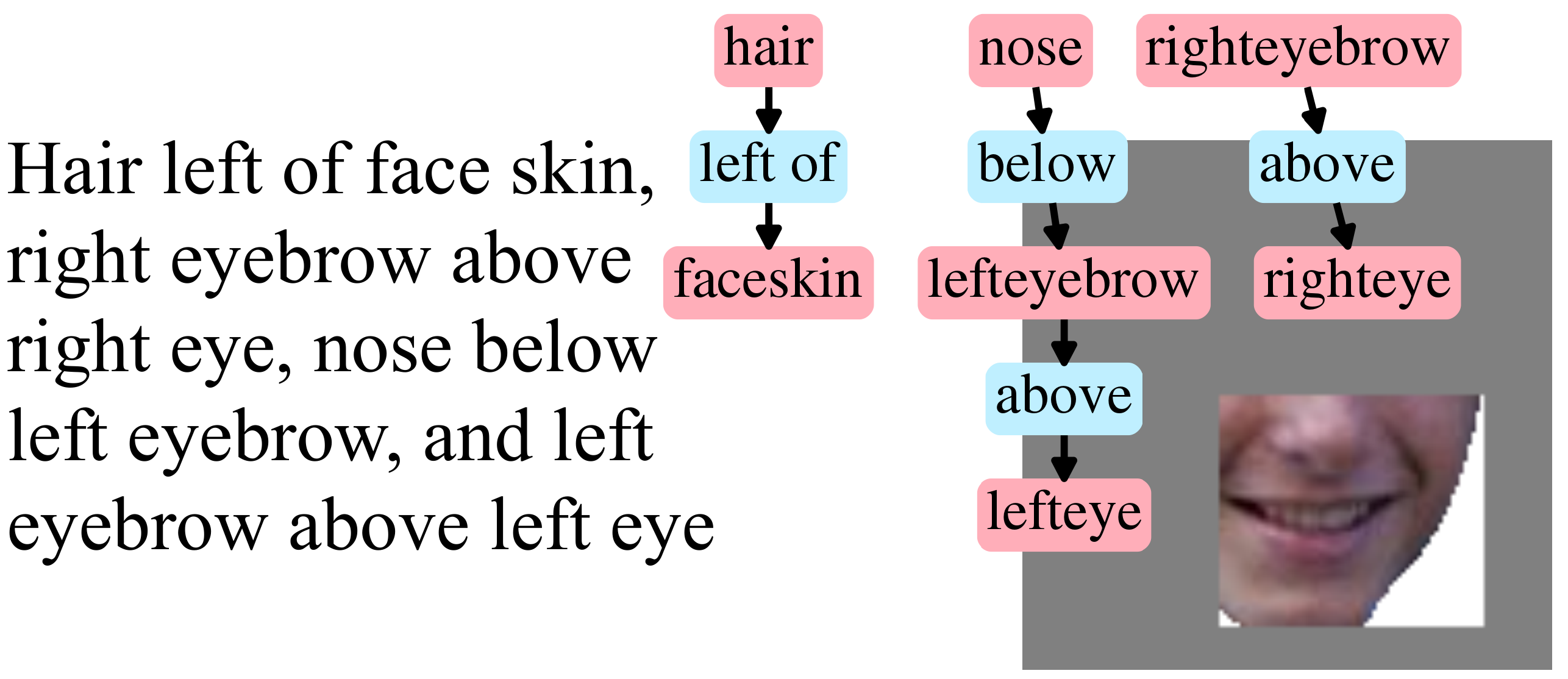} & 
\hspace{1pt}\vrule\hspace{1pt}
\includegraphics[width = .135\linewidth]{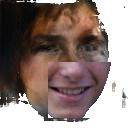} &
\includegraphics[width = .135\linewidth]{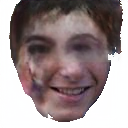} &
\includegraphics[width = .135\linewidth]{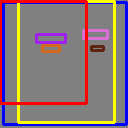} & 
\includegraphics[width = .135\linewidth]{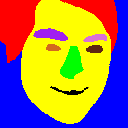} & 
\includegraphics[width = .135\linewidth]{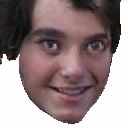} & \\

\includegraphics[width = .27\linewidth]{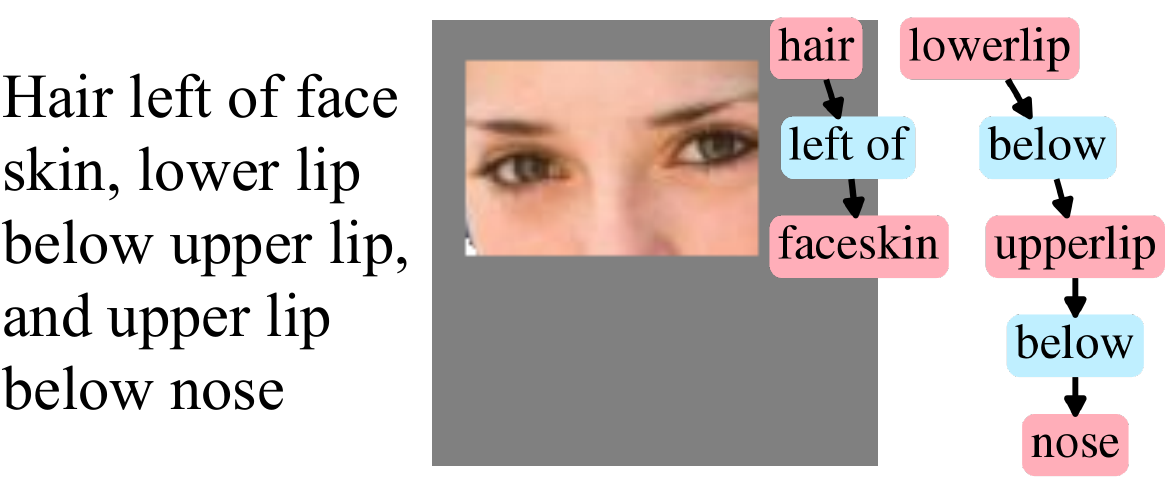} & 
\hspace{1pt}\vrule\hspace{1pt}
\includegraphics[width = .135\linewidth]{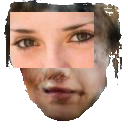} &
\includegraphics[width = .135\linewidth]{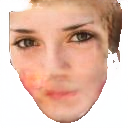} &
\includegraphics[width = .135\linewidth]{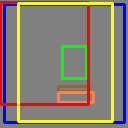} & 
\includegraphics[width = .135\linewidth]{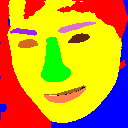} & 
\includegraphics[width = .135\linewidth]{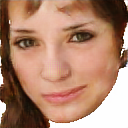} & \\

\includegraphics[width = .22\linewidth]{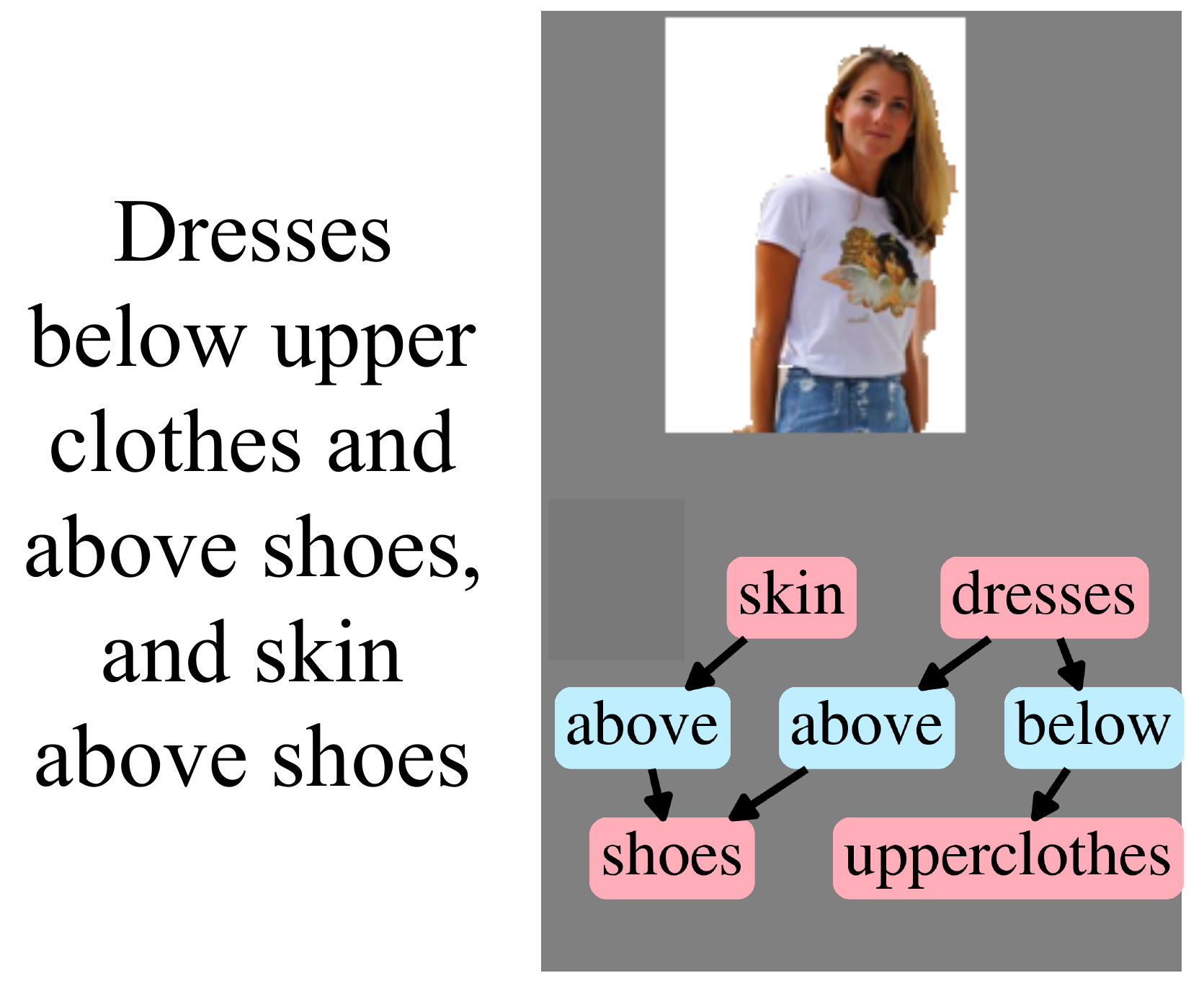} & 
\hspace{1pt}\vrule\hspace{1pt}
\includegraphics[width = .135\linewidth]{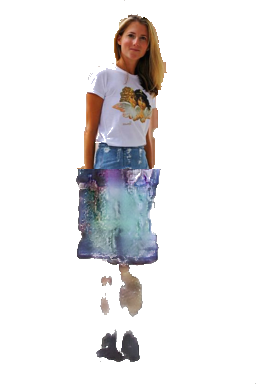} &
\includegraphics[width = .135\linewidth]{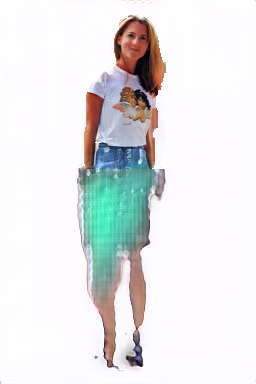} &
\includegraphics[width = .135\linewidth]{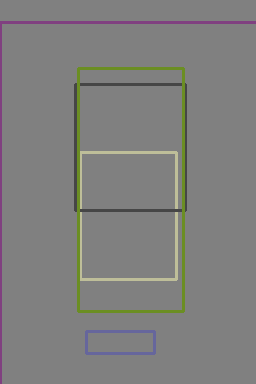} & 
\includegraphics[width = .135\linewidth]{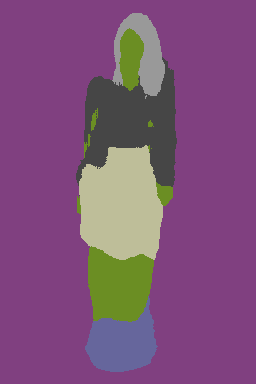} & 
\includegraphics[width = .135\linewidth]{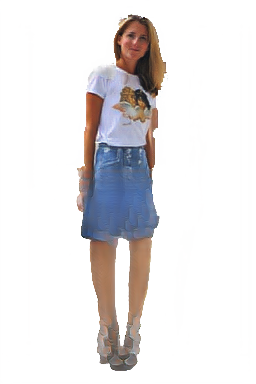} & \\

\includegraphics[width = .27\linewidth]{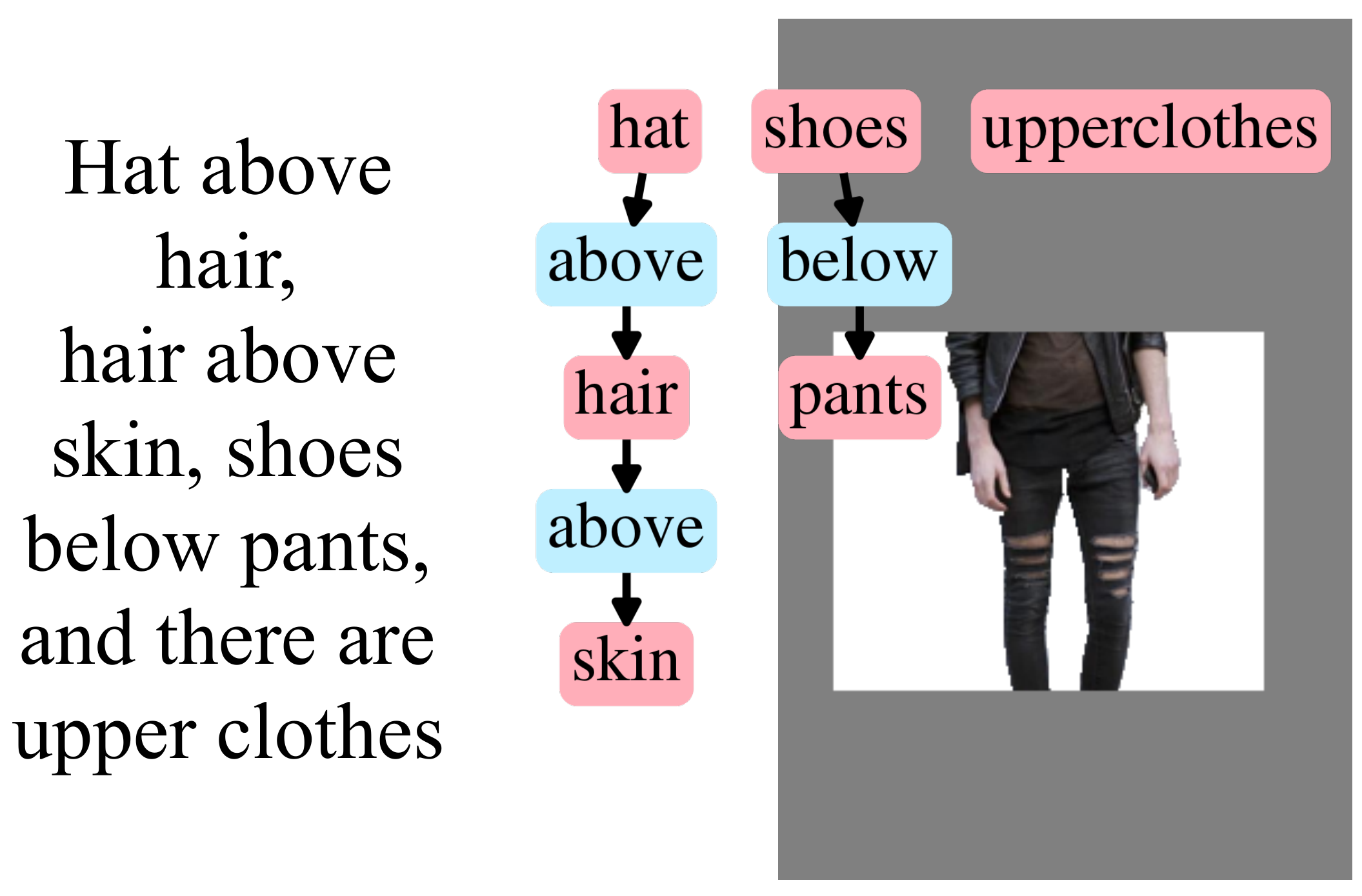} & 
\hspace{1pt}\vrule\hspace{1pt}
\includegraphics[width = .135\linewidth]{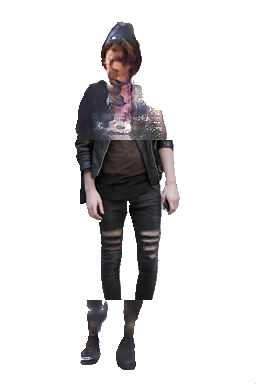} &
\includegraphics[width = .135\linewidth]{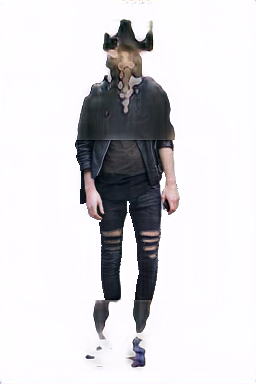} &
\includegraphics[width = .135\linewidth]{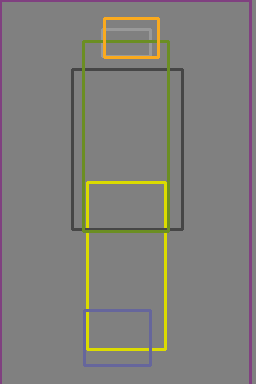} & 
\includegraphics[width = .135\linewidth]{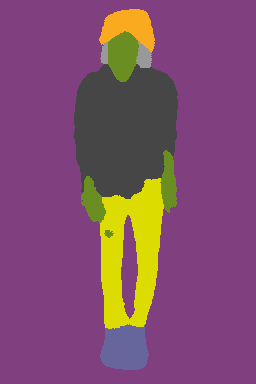} & 
\includegraphics[width = .135\linewidth]{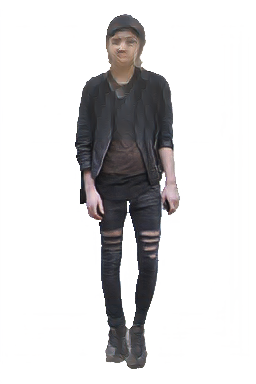} & \\

{(a) Input} & {(b) sg2im\_r} & {(c) sg2im\_c} & {(d) Bbox} & {(e) Segmentation}& {(f) Ours}\\

\end{tabular}
\caption{Visual comparisons between our method and baselines. Different colors in the layout (e) represent different object nodes.}
\label{fig:comparisons}
\end{figure*}

\vspace{0.5em}
\subsection{Experimental Setups}

\vspace{0.5em}
\noindent\textbf{Dataset.}~The Helen dataset~\cite{helendata-2012-interactive} consists of 2,330 face images with each face having 11 labels from~\cite{halenlabel-2013-exemplar} of main facial components.
The Clothing Co-Parsing (CCP) dataset~\cite{yang2014clothing} contains 1,004 images and corresponding label maps for 59 clothing items.
Since the label classes are highly unbalanced, we group similar labels (\eg, boots and wedges are both treated as shoes) and create a super label set of 9 clothing items: \{background, accessory, upper cloth, shoe, dress, hair, hat, pant, skin\}.
The experiments are conducted on these two datasets mainly because (i) face and human body are two types of object image of great interests, and (ii) compared with more complex scene datasets (\eg, COCO~\cite{lin2014microsoft}, Cityscapes~\cite{cordts2016cityscapes}) which only label the rough silhouette of objects, they contain more important detailed object parts which are useful for learning the controllable model.
%
%


The ground truth coordinates of the bounding-box of each label are computed by condiering the smallest and largest coordinate of all pixels with the same label as the top left and the bottom right.
Since both datasets do not provide annotated scene graphs, we construct the input scene graphs in a way  similar to~\cite{johnson-2018-sg2im} from the ground truth position of each label in the image, with each label as the node and one of the six spatial relationships $\{$left of, right of, above, below, inside, surrounding$\}$ as the edge.

During the training process, for each input image, we crop image patches of random size (around 15\%$\sim$25\% of the original image size) at random positions and train the network model to recover the original image.
We  fix the output size of extrapolated results which serves as a predefined canvas to mainly restrict the scale of objects in the results. 
The extrapolated iamge sizes of face and human body in our work are 128$\times$128 and 384$\times$256 pixels  respectively, which is 4$\sim$6 times bigger than the size of input patches.
For images in both datasets, we replace their original complex background, \ie pixels of the label $0$, with the clean white background to let the network focus on learning meaningful object parts.

\begin{figure}[t]
\begin{tabular}{c@{\hspace{0.005\linewidth}}c@{\hspace{0.005\linewidth}}c@{\hspace{0.005\linewidth}}c@{\hspace{0.005\linewidth}}c@{\hspace{0.005\linewidth}}c@{\hspace{0.005\linewidth}}c@{\hspace{0.005\linewidth}}c@{\hspace{0.005\linewidth}}c}

\includegraphics[width = .24\linewidth]{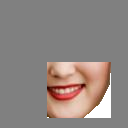} &
\includegraphics[width = .24\linewidth]{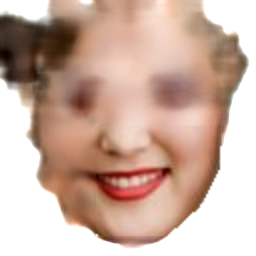} &
\includegraphics[width = .24\linewidth]{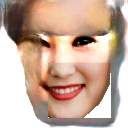} &
\includegraphics[width = .24\linewidth]{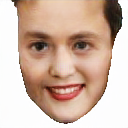} & \\

\includegraphics[width = .24\linewidth]{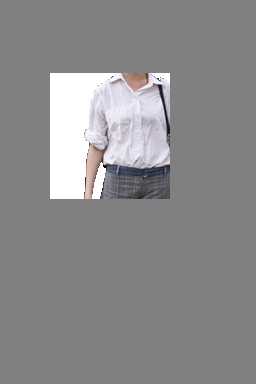} &
\includegraphics[width = .24\linewidth]{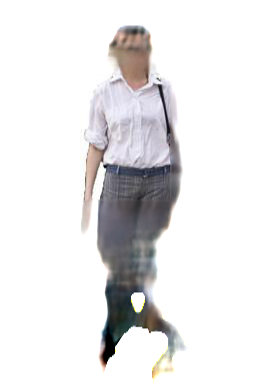} &
\includegraphics[width = .24\linewidth]{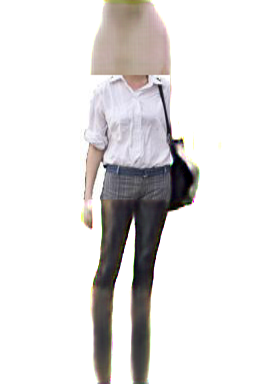} &
\includegraphics[width = .24\linewidth]{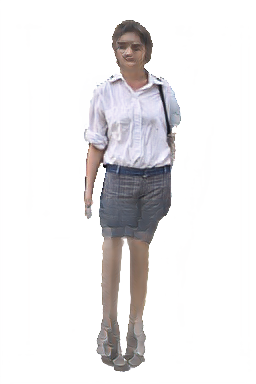}& \\

{Patch} & {GMCNN~\cite{wang-2018-image}}& {SRN~\cite{wang-2019-IE}} & {Ours}\\

\end{tabular}
\caption{Comparisons with non-text based inpainting/outpainting methods which directly generate the final output without taking the layout into account.}
\label{fig:compareother}
\vspace{-1em}
\end{figure}

\vspace{0.5em}
\noindent\textbf{Evaluated methods.}
~Since there exist no exact extrapolation methods that can handle the multimodal input of image patch and text (or scene graph), we compare with the following related work.
The \textbf{\emph{GMCNN}}~\cite{wang-2018-image} is the state-of-the-art image inpainting model. We adapt its original training objective from inpainting to outpainting pixels outside the patch boundary and keep the rest unchanged. As it does not support controls from text or scene graph, we train the model only based on the image patch using their released code\footnote{\footnotesize\url{https://github.com/shepnerd/inpainting_gmcnn}}.
The \textbf{\emph{SRN}}~\cite{wang-2019-IE} is the state-of-the-art model for unconditional image extrapolation. Similarly, the input to this model is an image patch only and we train the model using their code\footnote{\footnotesize\url{https://github.com/shepnerd/outpainting_srn}} on both Helen and CCP dataset. 
The \textbf{\emph{sg2im}}~\cite{johnson-2018-sg2im} is closely-related prominent method to synthesize image from scene graph. 
As it does not take the image patch as input, we simply copy and paste the image patch in the generated image. We denote this baseline as \textbf{\emph{sg2im\_r}} and train the model using the code from~\cite{johnson-2018-sg2im}\footnote{\footnotesize\url{https://github.com/google/sg2im}}.
The \textbf{\emph{sg2im\_c}} is an variant where we concatenate the image patch as additional input channels of their refinement network to train the model. The rest of the training condition is the same as in~\cite{johnson-2018-sg2im}. 
In addition, we also compare with a variant of the proposed method in terms of training strategy.
In contrast to the progressive training (pt) strategy used as default, we directly train all three stages from scratch and demoted this baseline as \textbf{\emph{Ours w/o pt}}.
For fair comparisons, we use the same set of cropped image patches in all methods.

\vspace{0.5em}
\noindent\textbf{Implementation details.}
~All models are trained on a single Tesla P100, using the Adam~\cite{kingma2014adam} optimizer with learning rate 1e\mbox{-}4 and batch size 16. 
The dimension of object and relationship embeddings is set as 128.
The ground truth bounding-box coordinates are normalized by the image height and width (in the range of [0, 1]).
Regarding the weight $\omega_{c_i}$ of each class in Eq.~\eqref{eq:cross_entropy}, for the Helen~\cite{helendata-2012-interactive} dataset, we set $\omega_{c_i}=1$ for classes of background and hair, and $\omega_{c_i}=10$ for other classes. 
For the CCP~\cite{yang2014clothing} dataset, we set $\omega_{c_i}=1$ for background, $\omega_{c_i}=10$ for classes of hair, hat and shoe, and $\omega_{c_i}=5$ for other classes.
During the separate training stages, we assume the input of each stage is already given from the ground truth data. For example, when training the Stage II alone, the input bounding-boxes are ground truth ones.
During the joint training stage, we connect all three stages (\ie, the output from previous stage will be the input of next stage) and the ground truth will be only used for computing losses in Eq.~\eqref{eq:gan_objective}.
The training process takes about one day to achieve convergence.

\begin{table*}[t]
  \caption{Quantitative evaluations on the Helen~\cite{helendata-2012-interactive} and CCP~\cite{yang2014clothing} dataset.
  }
  \vspace{0.5em}
  \label{table:quan_helen}
  \centering
  \begin{tabular}{cccccccc}
    \toprule
     & & sg2im\_r & sg2im\_c & GMCNN~\cite{wang-2018-image} & SRN~\cite{wang-2019-IE} & Ours w/o pt & Ours \\ 
    \midrule
    \multirow{2}{*}{Helen} & IS~$\uparrow$  & 1.73~$\pm$~0.14 & 1.42~$\pm$~0.08 & 1.40~$\pm$~0.11 & 1.48~$\pm$~0.12 & 1.45~$\pm$~0.11 & \textbf{1.82~$\pm$~0.16} \\ 
    & FID~$\downarrow$  & 120.28~$\pm$~1.51 & 70.02~$\pm$~1.53 & 71.28~$\pm$~1.22 & 67.69~$\pm$~1.63 & 62.34~$\pm$~1.27 & \textbf{49.21~$\pm$~1.92} \\ 
    \multirow{2}{*}{CCP} & IS~$\uparrow$  & 3.39~$\pm$~0.30 & 3.01~$\pm$~0.34 & 3.24~$\pm$~0.29 & 3.37~$\pm$~0.27 & 3.14~$\pm$~0.31 & \textbf{3.67~$\pm$0.33} \\
    & FID~$\downarrow$  & 141.82~$\pm$~1.13 & 119.77~$\pm$~0.19 & 95.88~$\pm$~1.49 & 86.85~$\pm$~1.22 & 97.24~$\pm$~0.78 & \textbf{68.64~$\pm$~0.17} \\ 
    
    \bottomrule
  \end{tabular}
\end{table*}

%
%
%
%

%
%
%
%
%


\subsection{Qualitative Comparison}\label{sec:qualitative}
~Figure~\ref{fig:comparisons} and Figure~\ref{fig:compareother} show the visual comparisons between the proposed method and baselines, where the former includes conditional extrapolation and the latter contains unconditional extrapolation baselines. 
Given the scene graph and conditional image patch in in Figure~\ref{fig:compareother}(a), our method generates more visually appealing and realistic results (f) than the baseline methods (b)-(c).
We also show the intermediate outputs of each stage in our model, \ie, the coarse bounding-box layout in (d) and the segmentation layout in (e). 
Figure~\ref{fig:compareother} shows that the inpainting and outpainting algorithms, which uses no text inputs, are missing the majority of pixels about detailed object parts (\eg, the thin eyebrow and small head). 
Overall, our model generate sharper and more realistic results. 
%
%
%

\begin{figure}[t]
\centering
\begin{tabular}{c@{\hspace{0.005\linewidth}}c@{\hspace{0.005\linewidth}}c@{\hspace{0.005\linewidth}}c@{\hspace{0.005\linewidth}}c@{\hspace{0.005\linewidth}}c@{\hspace{0.005\linewidth}}c@{\hspace{0.005\linewidth}}c@{\hspace{0.005\linewidth}}c}

\includegraphics[width = .18\linewidth]{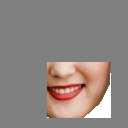} &

\hspace{1pt}\vrule\hspace{1pt}

\includegraphics[width = .36\linewidth]{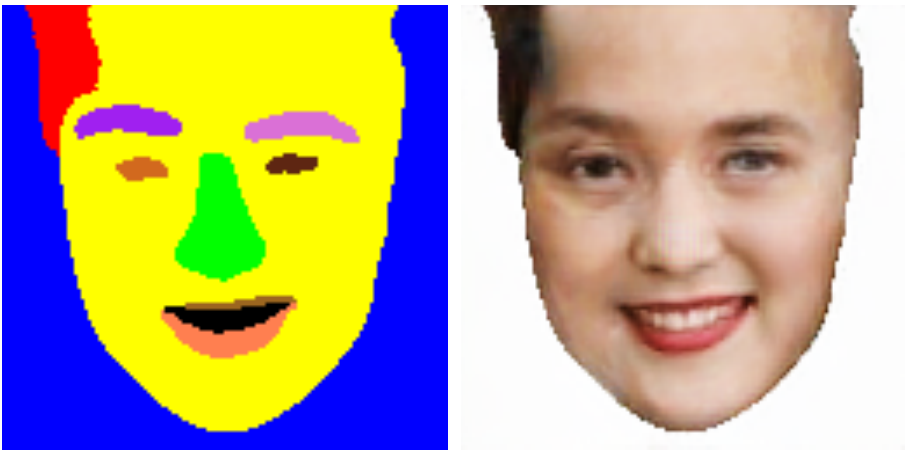} &

\hspace{1pt}\vrule\hspace{1pt}

\includegraphics[width = .36\linewidth]{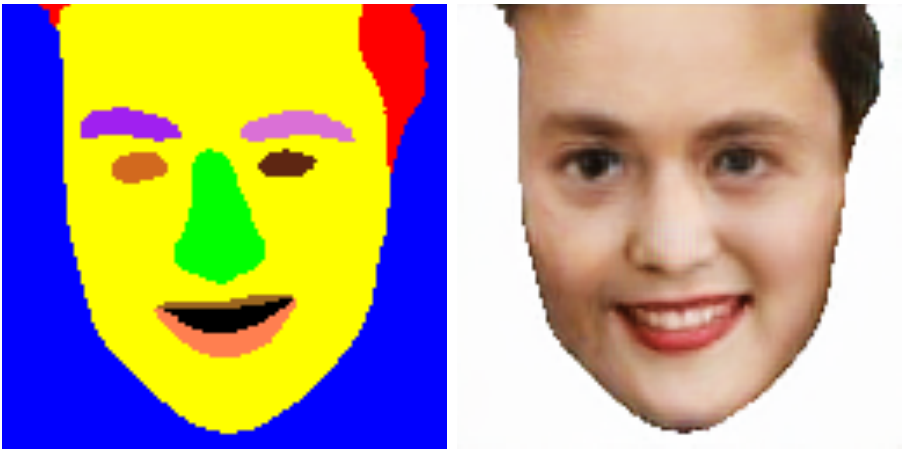} & \\



{Patch} & {Hair on left} & {Hair on right} \\ 

\includegraphics[width = .18\linewidth]{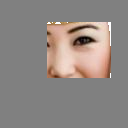} &

\hspace{1pt}\vrule\hspace{1pt}

\includegraphics[width = .36\linewidth]{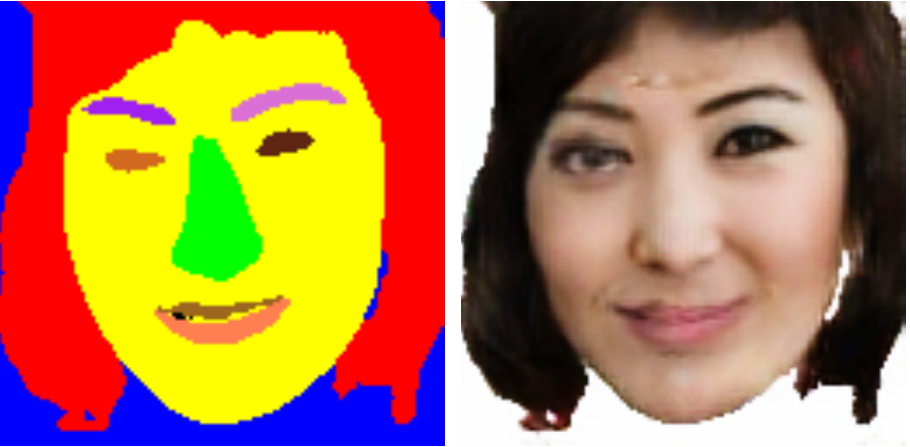} & 
\includegraphics[width = .36\linewidth]{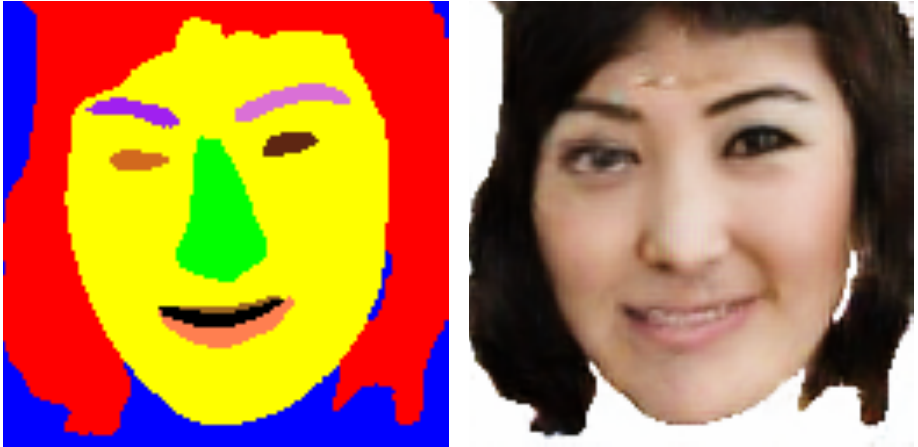} & \\

{Patch} & {No inner mouth} & {With inner mouth} \\

\includegraphics[width = .18\linewidth]{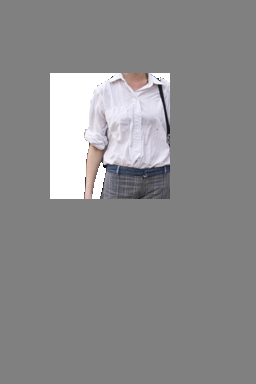} &

\hspace{1pt}\vrule\hspace{1pt}

\includegraphics[width = .36\linewidth]{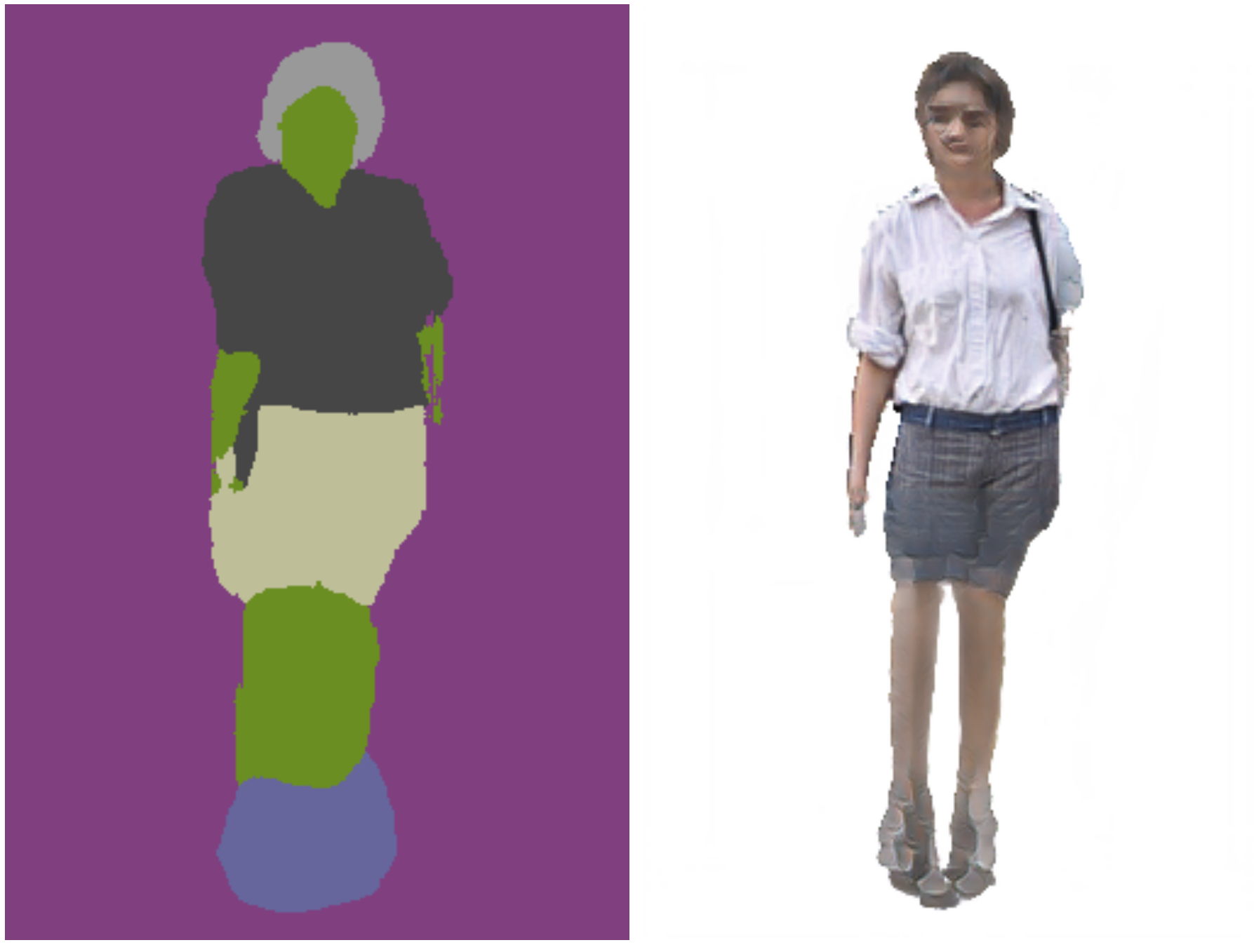} &

\hspace{1pt}\vrule\hspace{1pt}

\includegraphics[width = .36\linewidth]{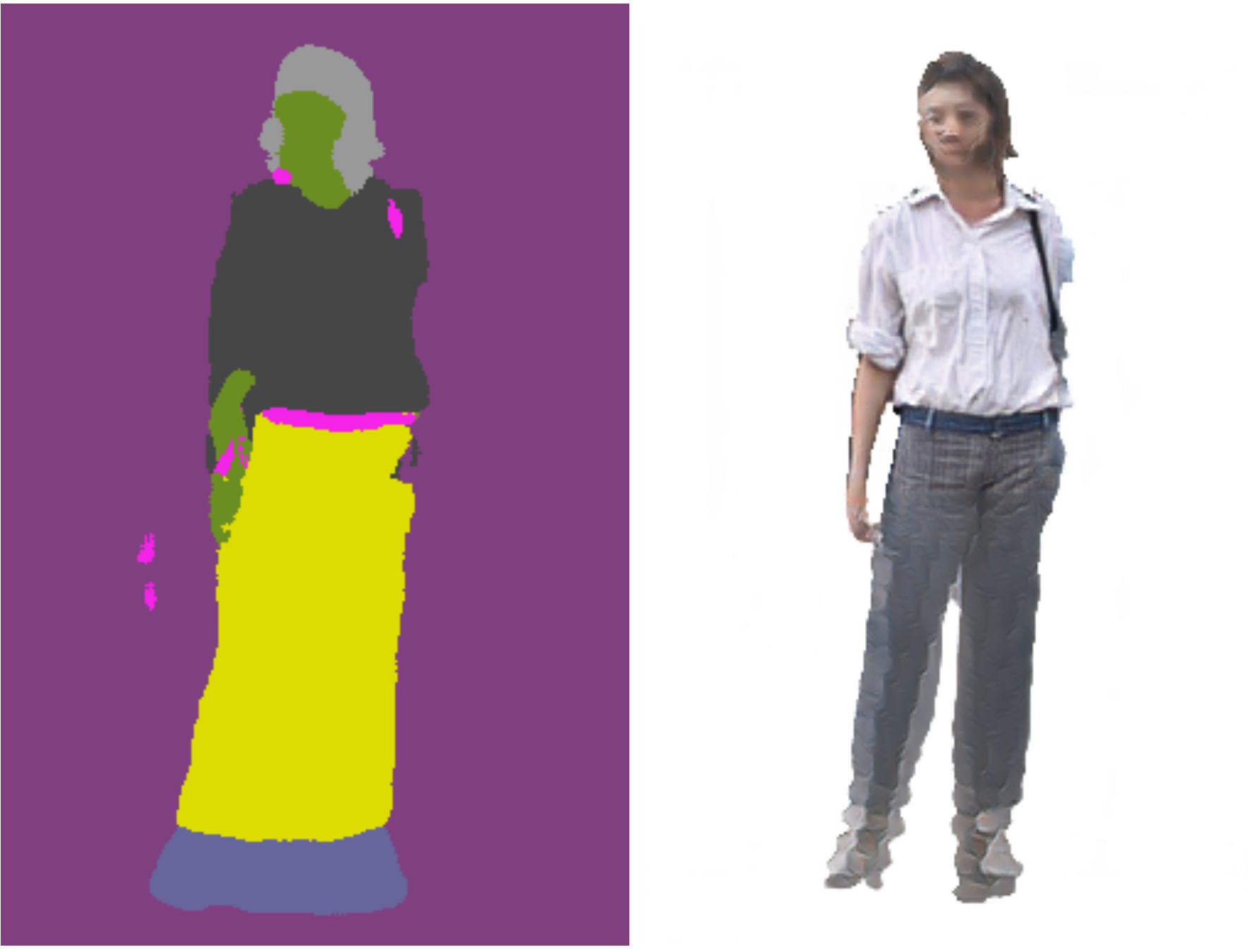} & \\

{Patch} & {Wearing dress} & {Wearing pants} \\

\includegraphics[width = .18\linewidth]{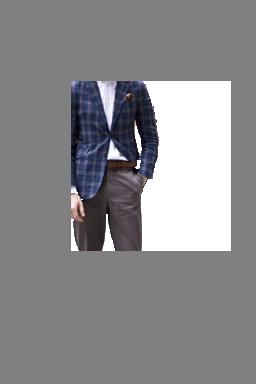} & 

\hspace{1pt}\vrule\hspace{1pt}

\includegraphics[width = .36\linewidth]{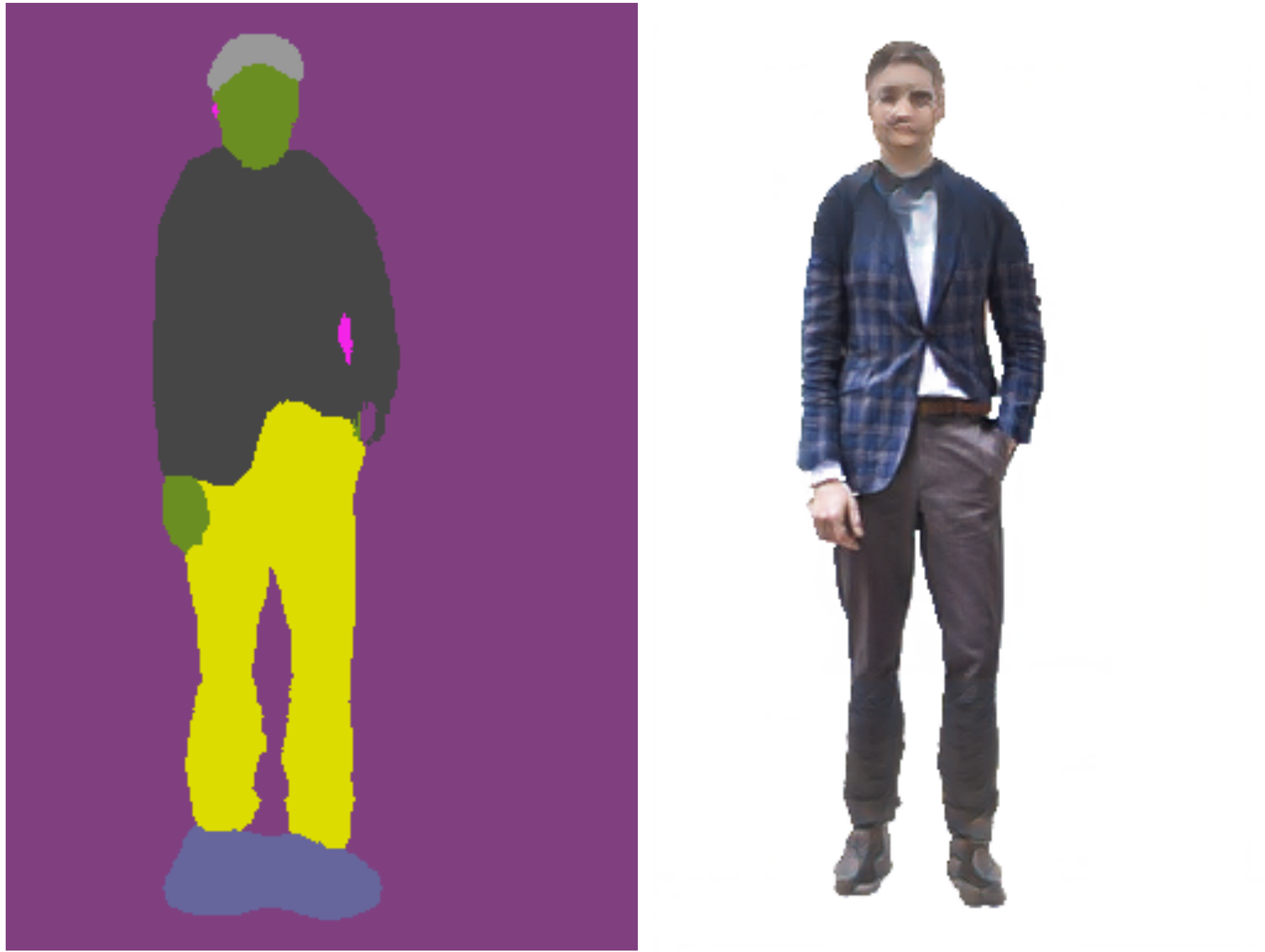} & 

\hspace{1pt}\vrule\hspace{1pt}

\includegraphics[width = .36\linewidth]{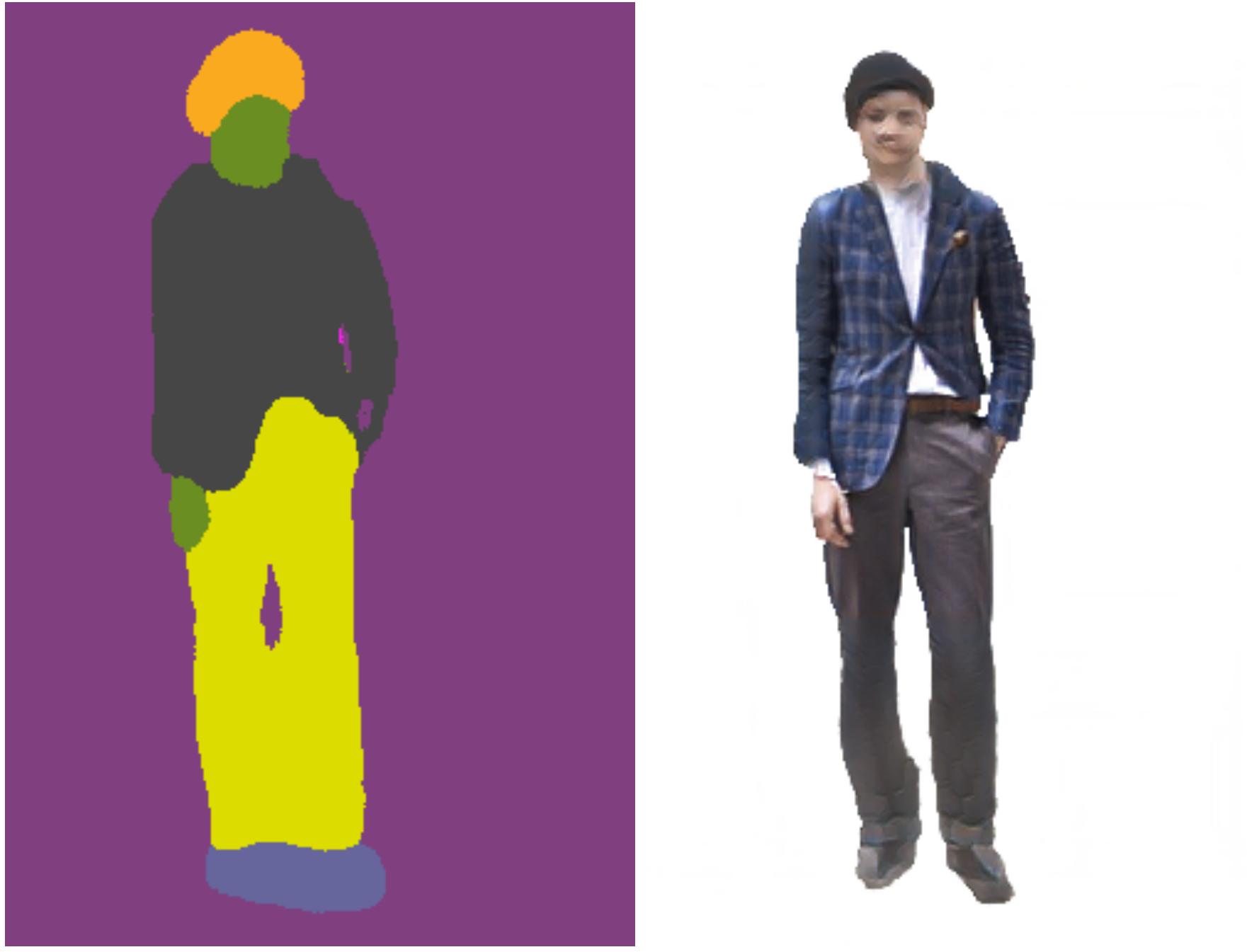} & \\

{Patch} & {No hat} & {Wearing hat} \\

\end{tabular}
\caption{Diverse results by manipulating the scene graphs.}
\label{fig:control}
\vspace{-1em}
\end{figure}

\subsection{Quantitative results}
~We first evaluate the realism of the extrapolated results, \ie measuring how close the distribution of results is to that of the real data. We use two common metrics for general image generation tasks: Inception Score (IS)~\cite{salimans2016-improved} and Fr\'echet Inception Distance (FID)~\cite{heusel2017-fid}. We randomly crop patches on images in the test set and compute the metrics over 3,000 outputs of each model. Note that these metrics favor realistic and reasonable images completely neglecting the input texts (scene graph) and hence cannot evaluate the controllable setting.
%
%
%
%
The evaluation results, listed in Table~\ref{table:quan_helen}, shows that the proposed method achieves higher IS and lower FID scores than all baselines across the datasets. 
%

%
%

We conduct user studies to analyse human perceptual preference towards different methods.
In addition to visual quality, we also concern the relevance of generated images to the input scene graph. 
Thus here we only compare our model with the baseline methods sg2im\_r and sg2im\_c that are able to control the extrapolation by scene graphs. 
We prepare extrapolated images for 20 (10 from Helen~\cite{helendata-2012-interactive} and 10 from CCP~\cite{yang2014clothing}) pairs of scene graphs and patches. 
For each subject, we randomly select 10 pairs to evaluate and display the extrapolated results side-by-side in random order. 
%
%
Each subject is asked to vote the single best generated image that are (i) relevant to the given scene graph and (ii) realistic.
%
We collect 480 votes from 48 expert participants who are not involved in the project. 
The user study is double-blind, \ie, our results are shown unlabeled in randomized order and the identities of the participants are not disclosed.
The user study results show that the proposed method receives the most votes, significantly higher than others. 
These results substantiate that our model is able to generate controllable image content that are more semantic relevant to the input scene graph.

It is relatively difficult to evaluate the layout generated in Stage I and II because there is no unique ground truth for an input, especially under a controllable setting where one conditioned patch could be extrapolated into different results with different scene graph inputs. 
Therefore we mainly focus on the evaluation of final results using the IS/FID metric and user studies, where humans can examine relevance between the final image and the given text.

%
%
%
%
%

\begin{table}[t]
  \caption{ User preference towards different methods (\%).} 
  \vspace{0.5em}
  \label{table:user_study}
  \centering
  \begin{tabular}{cccc}
    \toprule
     & sg2im\_r & sg2im\_c & Ours \\
    \midrule
    Preference (\%)~$\uparrow$  & 2.65 & 5.01 & \textbf{92.34} \\ 
    \bottomrule
  \end{tabular}
\end{table}

\begin{figure}[t]
\begin{tabular}{c@{\hspace{0.005\linewidth}}c@{\hspace{0.005\linewidth}}c@{\hspace{0.005\linewidth}}c@{\hspace{0.005\linewidth}}c@{\hspace{0.005\linewidth}}c@{\hspace{0.005\linewidth}}c@{\hspace{0.005\linewidth}}c@{\hspace{0.005\linewidth}}c}

\includegraphics[width = .5\linewidth]{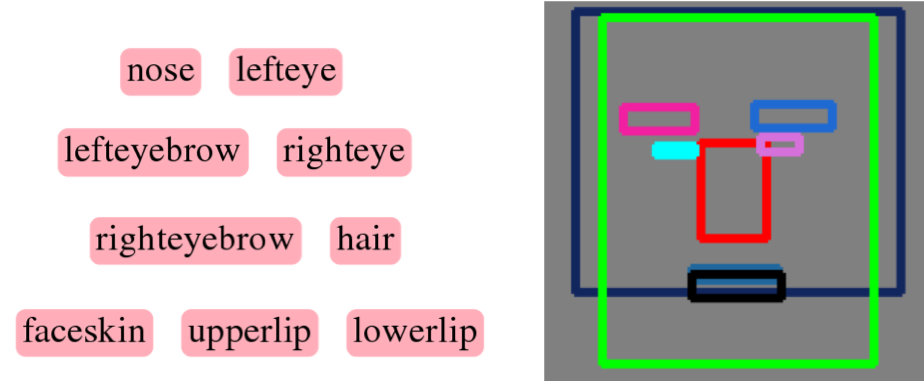} & 

\hspace{1pt}\vrule\hspace{1pt} 

\includegraphics[width = .21\linewidth]{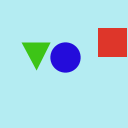} & 
\includegraphics[width = .21\linewidth]{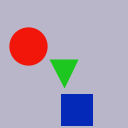} \\

\end{tabular}

\caption{Left: the generated layout without relationships in scene graph, trained on real data (faces) where there are priors over certain object parts. Right: two examples in the synthetic shape dataset~\cite{xue2018visual} where the relational positions between objects are completely random, without priors.}
\label{fig:prior}
\end{figure}

\subsection{Ablation studies on controllability} 

A unique property of conditional image extrapolation is being able to control image extrapolation with different text inputs. 
%
%
Figure~\ref{fig:control} shows different extrapolated results of our model from the same image patch for different inputs.
We randomly change the node or the relation of a given scene graph at a time.
The results show that our extrapolation model follows the control signals specified in the scene graph and generate images that align well with the conditional image patch.
More results can be found in the supplementary material.
These results tests our model is able to control extrapolation based on texts and images.

\vspace{0.5em}
\noindent\textbf{Priors vs. Controls.}~We observe that for real object data, there exist strong priors over certain object parts, \eg, lips are always under noses in faces or sky is always above other objects. 
During training, network models will bias towards the prior and ignore the input control signal. 
The prior inherently exists in our natural world and every real dataset, simple or complex, big or small.
%
To demonstrate this, we use a simple experiment by removing all relationships in the scene graph.
Figure~\ref{fig:prior} shows that with object nodes only, the model is still able to generate a reasonable layout. However, we do not want to completely lose the controls over the extrapolation.
As shown in Figure~\ref{fig:comparisons}, we can still control several items like the position of hair, pant or dress, and with or without hat.

\begin{figure}[t]
\centering
\begin{tabular}{c@{\hspace{0.005\linewidth}}c@{\hspace{0.005\linewidth}}c@{\hspace{0.005\linewidth}}c@{\hspace{0.005\linewidth}}c@{\hspace{0.005\linewidth}}c@{\hspace{0.005\linewidth}}c@{\hspace{0.005\linewidth}}c@{\hspace{0.005\linewidth}}c}

\includegraphics[width = .24\linewidth]{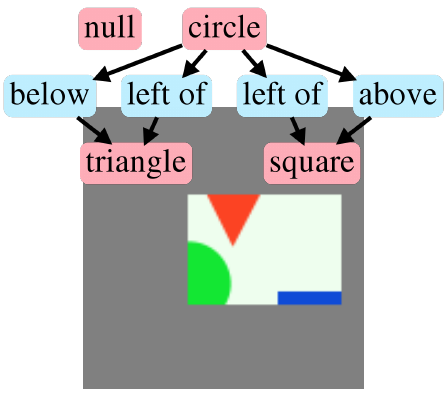} & 
\includegraphics[width = .24\linewidth]{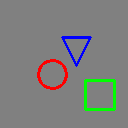} & 
\includegraphics[width = .24\linewidth]{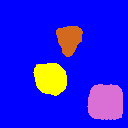} & 
\includegraphics[width = .24\linewidth]{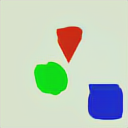} & \\

{Input} & {Bbox} & {Segmentation} & {Our result} \\

\end{tabular}
\caption{Layout generation on synthetic shape dataset (the node \emph{null} means the background).}
\label{fig:shape1}
\vspace{-0.5em}
\end{figure}

\begin{figure}[t]
\centering
\begin{tabular}{c@{\hspace{0.005\linewidth}}c@{\hspace{0.005\linewidth}}c@{\hspace{0.005\linewidth}}c@{\hspace{0.005\linewidth}}c@{\hspace{0.005\linewidth}}c@{\hspace{0.005\linewidth}}c@{\hspace{0.005\linewidth}}c@{\hspace{0.005\linewidth}}c}

\includegraphics[width = .15\linewidth]{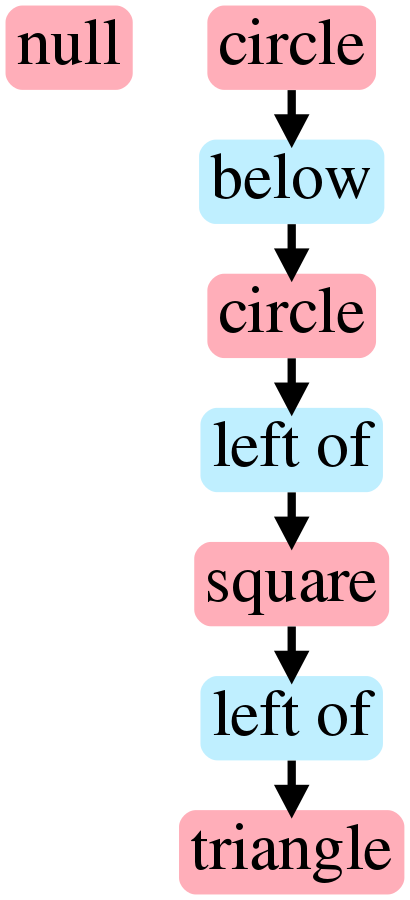} & 
\includegraphics[width = .31\linewidth]{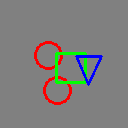} & 
\includegraphics[width = .18\linewidth]{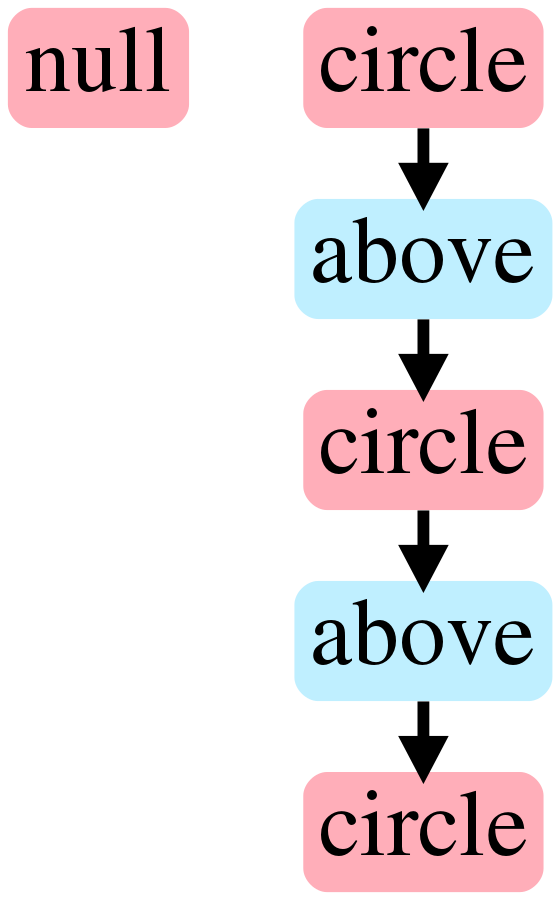} & 
\includegraphics[width = .31\linewidth]{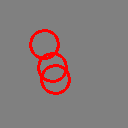} & \\

\includegraphics[width = .15\linewidth]{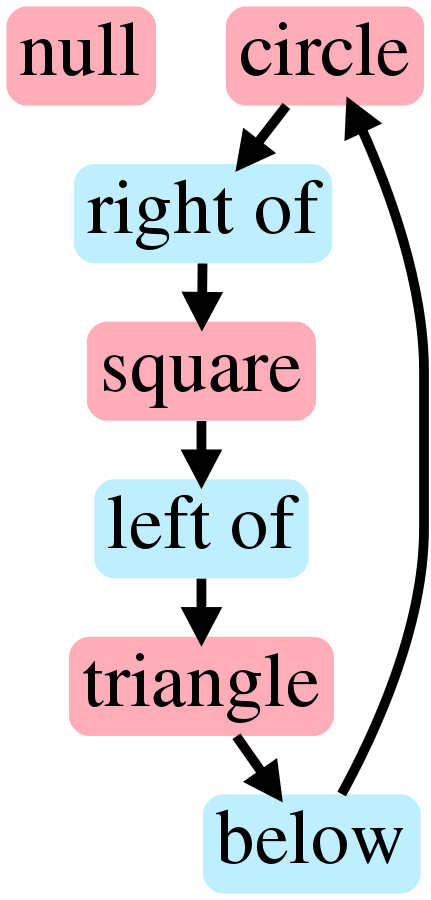} & 
\includegraphics[width = .31\linewidth]{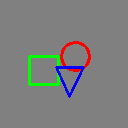} & 
\includegraphics[width = .15\linewidth]{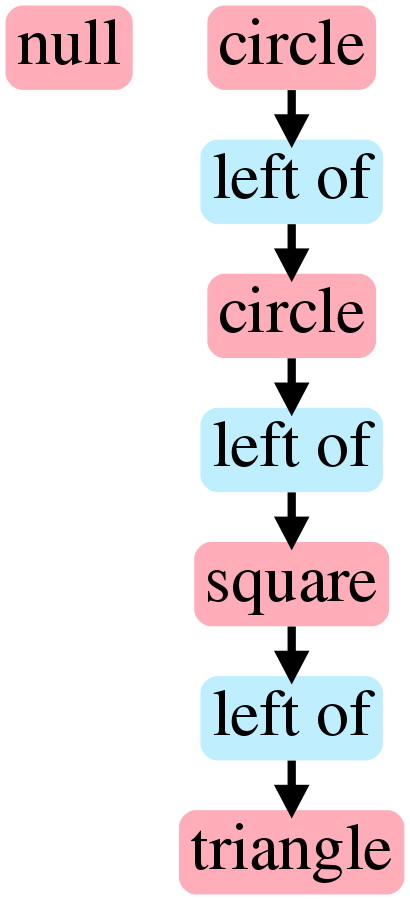} & 
\includegraphics[width = .31\linewidth]{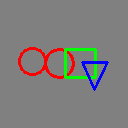} & \\

\includegraphics[width = .15\linewidth]{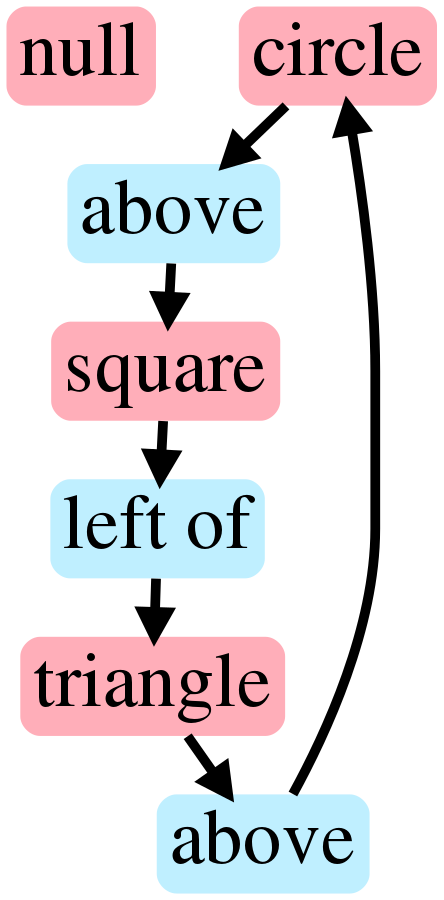} & 
\includegraphics[width = .31\linewidth]{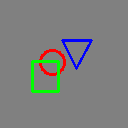} &  
\includegraphics[width = .15\linewidth]{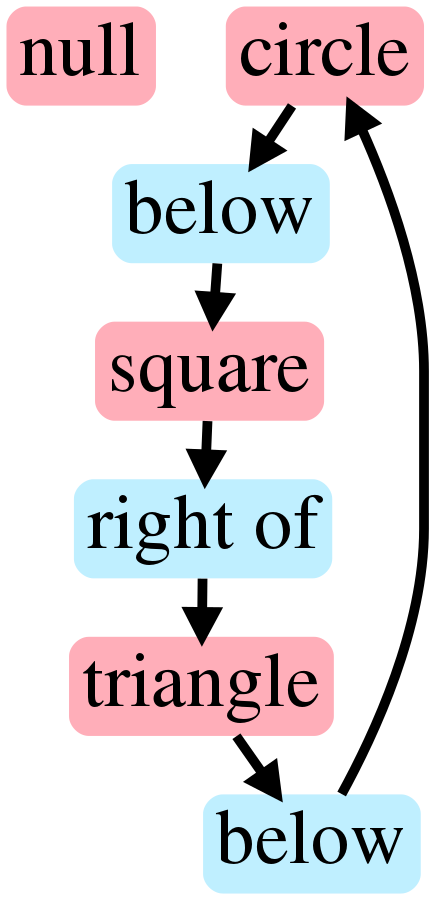} & 
\includegraphics[width = .31\linewidth]{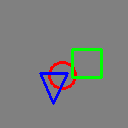} & \\



\end{tabular}
\caption{Flexible controls from scene graphs at Stage I.}
\label{fig:shape}
\vspace{-1em}
\end{figure}

To further validate the effective controls by the scene graph, we conduct experiments on a synthetic dataset of 2D shapes~\cite{xue2018visual}. 
Each image in~\cite{xue2018visual} contains three types of objects (circles, squares, and triangles). They are randomly positioned to reduce the prior information (two examples are presented on right of Figure~\ref{fig:prior}).
We show an example of our outpainting results in Figure~\ref{fig:shape1}. 
By manipulating the scene graph, our model is able to generate diverse bounding-box layouts as shown in Figure~\ref{fig:shape}, where each generated layout correctly reflects the object and relationship information in the scene graph.
Note that each image in the original shape dataset contains one circle, one square and one triangle only. Our model generates more combinations of three categories (\eg, multiple circles) through controlling the scene graph. 
This can be potentially used for graphic layout design.

It is also worth noting that although the scene graph provides control signals, we find it is insufficient to model rare objects or relationships. 
For example, it is unlikely to generate four left eyebrows if there are four left eyebrows in the scene graph. This is expected because there exist no such cases in the training dataset.
From the experimental results on both real and synthetic datasets, we conclude that the controllability of scene graphs can be flexible but will be constrained, at least to some extent, by the data prior. 
%
%
%

\section{Conclusion}
In this work, we propose a generative network to extrapolate new content outside the image boundaries.
Unlike image extrapolation, the studied extrapolation is controlled by a structured text (modeled as a scene graph) indicating what and where to generate for the unknown region. 
To realize controllable extrapolation, we decompose the learning process into three stages and introduced two important sub-tasks, of generating layouts from coarse to fine, to facilitate the training.
%
%
%
Based on this multi-stage model, we use a curriculum learning strategy for effective model training.
%
%
Both qualitative and quantitative results show that the proposed model performs favorably against the evaluated methods and is able to generate more controllable extrapolated results.
%


{\small
\bibliographystyle{ieee_fullname}
\bibliography{egbib}
}

\end{document}